\documentclass[letterpaper, 10 pt, conference]{ieeeconf}  

\usepackage{amsmath,amssymb}
\usepackage[overload]{empheq}
\usepackage{graphics}
\usepackage[export]{adjustbox}
\usepackage{epsfig}
\usepackage{booktabs}
\usepackage{cite}
\usepackage{url}
\usepackage{subfig}
\usepackage{multirow}
\usepackage{siunitx}
\usepackage{eurosym}
\usepackage{enumerate}

\usepackage{caption}

\newcommand*\xbar[1]{%
  \hbox{%
    \vbox{%
      \hrule height 0.5pt 
      \kern0.5ex
      \hbox{%
        \kern-0.1em
        \ensuremath{#1}%
        \kern-0.1em
      }%
    }%
  }%
}

\IEEEoverridecommandlockouts                              

\title{\LARGE \bfseries Multi-Object Tracking with Camera-LiDAR Fusion for Autonomous Driving}

\author{Riccardo Pieroni, Simone Specchia, Matteo Corno$^*$, Sergio Matteo Savaresi
\thanks{The authors are with the Dipartimento di Elettronica, Informazione e Bioingegneria, Politecnico di Milano, Piazza Leonardo da Vinci 32, 20133 Milan, Italy. Email: \{{\tt\small riccardo.pieroni, simone.specchia,  matteo.corno, sergio.savaresi\}@polimi.it.} }
\thanks{$^*$ Corresponding author}}%

\begin{document}

\maketitle
\thispagestyle{empty}
\pagestyle{empty}

\begin{abstract}
This paper presents a novel multi-modal Multi-Object Tracking (MOT) algorithm for self-driving cars that combines camera and LiDAR data. 
Camera frames are processed with a state-of-the-art 3D object detector, whereas classical clustering techniques are used to process LiDAR observations.
The proposed MOT algorithm comprises a three-step association process, an Extended Kalman filter for estimating the motion of each detected dynamic obstacle, and a track management phase. The EKF motion model requires the current measured relative position and orientation of the observed object and the longitudinal and angular velocities of the ego vehicle as inputs. Unlike most state-of-the-art multi-modal MOT approaches, the proposed algorithm does not rely on maps or knowledge of the ego global pose. Moreover, it uses a 3D detector exclusively for cameras and is agnostic to the type of LiDAR sensor used. The algorithm is validated both in simulation and with real-world data,  with satisfactory results. 
\end{abstract}

\section{Introduction}
The advent of self-driving vehicles may revolutionize transportation.  An essential task for self-driving cars and autonomous vehicles is to detect and avoid obstacles. Planning obstacle avoidance maneuvers requires an estimate and prediction of the motion of other agents present in the scene. The problem of tracking multiple objects simultaneously is known in the scientific literature as Multi-Object Tracking (MOT).

The most common approaches in multi-object tracking exploit measurements collected from one or multiple sensors, which need to be linked to existing \textit{tracks} (\textit{i.e.} the detected moving objects) or utilized to create new ones. Existing MOT methods are classified either as single-modality-based or multi-modality-based methods. In the single-modality context, a single sensor type, such as a camera or LiDAR, is employed to gather data and predict object trajectories \cite{b1,b2,b3,b5,b6,b7,b11}. On the other hand, multi-modal methods combine inputs from different sensors, exploiting their different attributes to improve tracking accuracy and effectiveness. These approaches usually combine LiDAR and camera observations \cite{b4,b8,b9,b10}. This is because LiDAR sensors provide precise measurements of objects location, while cameras more effectively classify the type of road users in the scene (\textit{e.g.} pedestrians, cyclists, vehicles, ecc.). Fusing these sources improves tracking accuracy by filtering out static objects and focusing on dynamic entities. 

Regardless of the mode used, whether single or multi-modal, state-of-the-art MOT algorithms comprise three phases: tracks-measures association, tracks prediction/correction, and tracks management. The overall performance of a MOT algorithm relies on the quality of these three phases and on the performance of the detection algorithms employed.

In the existing literature, most MOT methods have been developed in a tracking-by-detection framework. This means that the multi-object tracking algorithm is developed based on the output of an object detection algorithm. State-of-the-art multi-object tracking uses 3D detectors to estimate the position, orientation, and type of objects in the scene \cite{b2,b4,b8,b9,b10}. These methods propose algorithms for the tracks-measures association and tracks management phases. 

Advancements in neural networks have also given rise to different MOT algorithms, including joint detection and tracking methodologies \cite{b1,b3,b6,b7}. These algorithms unify the tasks of object detection and tracking within a single neural network framework. These studies assume that the detection phase is the most complex one, simplifying the prediction, correction and association steps. However, many studies demonstrate that the tracks prediction/correction step and the association phase greatly influence the performance of the MOT. \cite{b2} describes how performing association and correction in the 2D image plane improves multi-object tracking performance for camera-only MOTs. This is because 3D measurements from a monocular 3D detector tend to have highly correlated errors over time, which cannot be accounted for by the Kalman filter, which does not allow biased or correlated errors over time. In contrast, \cite{b9} shows that, for multi-modal LiDAR-camera MOT, performing the association in the 3D Cartesian plane results in a better MOT performance, as objects are more susceptible to occlusion in the 2D plane. 

The method used for prediction influences the association step, since the better the prediction of the trajectories, the better the association with the new measures. There are different approaches to predict the motion of objects: \cite{b2} uses an Extended Kalman Filter (EKF) for each dynamic object, exploiting a kinematic single-track model for vehicles and cyclists, and a constant turn rate and velocity (CTRV) model for pedestrians. This approach requires knowledge of the ego vehicle's global position; \cite{b1} uses a prediction LSTM (P-LSTM), which models the dynamic object position in 3D coordinates by predicting the object velocity from previously updated velocities and position; \cite{b7} learns to predict a two-dimensional velocity estimation for each detected object as an additional regression output, using only the center of each object's bounding box ($x,y$). To obtain the velocity estimate, two map-views are required: the current and previous time-steps; \cite{b4} starts from the measure given by \cite{b7} and refines it with a Kalman Filter, assuming constant linear and angular velocity. This model requires the detector to provide a velocity estimate.

In this paper, we propose a multi-modal multi-object tracking algorithm based on the fusion of camera and LiDAR observations, with the following original contributions:
\begin{itemize}
    \item Our approach employs an EKF for tracking each object and a novel motion model that estimates the absolute longitudinal and angular velocities of an object. The EKF motion model only requires the current measured relative position and orientation of the observed object and the longitudinal and angular velocities of the ego vehicle as inputs, without relying on maps or knowledge of the ego global pose. 
    \item The extended Kalman filter accepts a vector of measurements that can have varying dimensions. Specifically, the measurements supplied by LiDARs are processed and used to correct the position ($x,y$), while the orientation ($\psi$) is corrected by exploiting the measurements provided by a camera. Depending on which measurements are available at each time instant, the EKF corrects either all three values ($x,y,\psi$) or a subset of them.
    \item The proposed approach uses the 3D detector \cite{b12} exclusively for cameras, unlike most multi-modal approaches in the literature, which use a 3D detector for both LiDAR and camera measurements. LiDAR centroids are calculated using an Euclidean clustering algorithm, which accelerates the execution of the entire algorithmic pipeline and ensures fast execution. Besides, detectors based on LiDAR often strongly rely on the pointcloud's structure which significantly varies depending on the type of LiDAR used. The proposed approach is instead agnostic to the type of LiDAR sensor used.
\end{itemize}

\section{Multi-Object Tracking Algorithm}
The goal of the proposed multi target tracking algorithm is to detect and track  dynamic (\emph{i.e.} moving) obstacles in the scene, estimating their position and linear/angular velocities.
In addition to the motion estimation, the algorithm associates to each obstacle a class, providing additional information to the planning layer. 

\begin{figure}[h] 
    \centering 
    \includegraphics[width=0.45\textwidth]{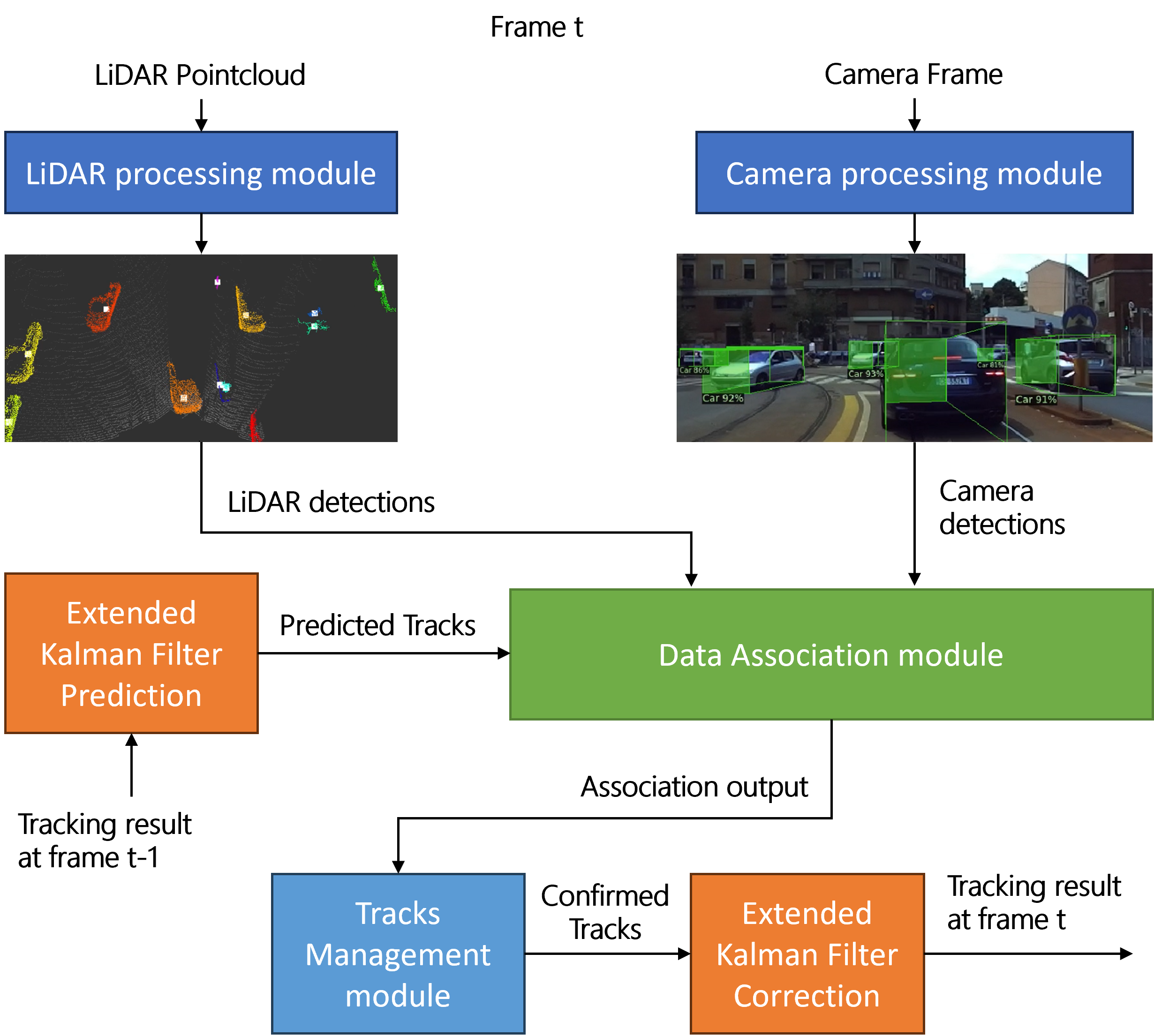}
    \caption{Schematic representation of the proposed MOT algorithm.} 
    \label{fig:algorithm} 
\end{figure}

The proposed method, illustrated in Figure \ref{fig:algorithm}, consists of four primary algorithmic blocks:

\begin{itemize}
    \item \textbf{Camera and LiDAR processing modules:} these modules process the raw input coming respectively from the camera and LiDAR sensors, returning  a set of observations of  obstacles in the scene. 
    \item \textbf{Data association:} at each time instant, the sensor processing modules will return multiple measurements. This step is necessary to group together camera and LiDAR observations generated by the same object and to possibly associate them to a pre-existing track representing the dynamic obstacle.
    \item \textbf{Extended Kalman Filter:} the EKF estimates the motion of the identified tracks, based on an internal motion model and exploiting the LiDAR and camera measurements.
    \item \textbf{Tracks management:} this step deals with initializing or deleting tracks, considering both the output of the association step and the time history of existing tracks.
\end{itemize}

\subsection{Camera processing}
\begin{figure}[h] 
    \centering 
    \includegraphics[width=0.45\textwidth]{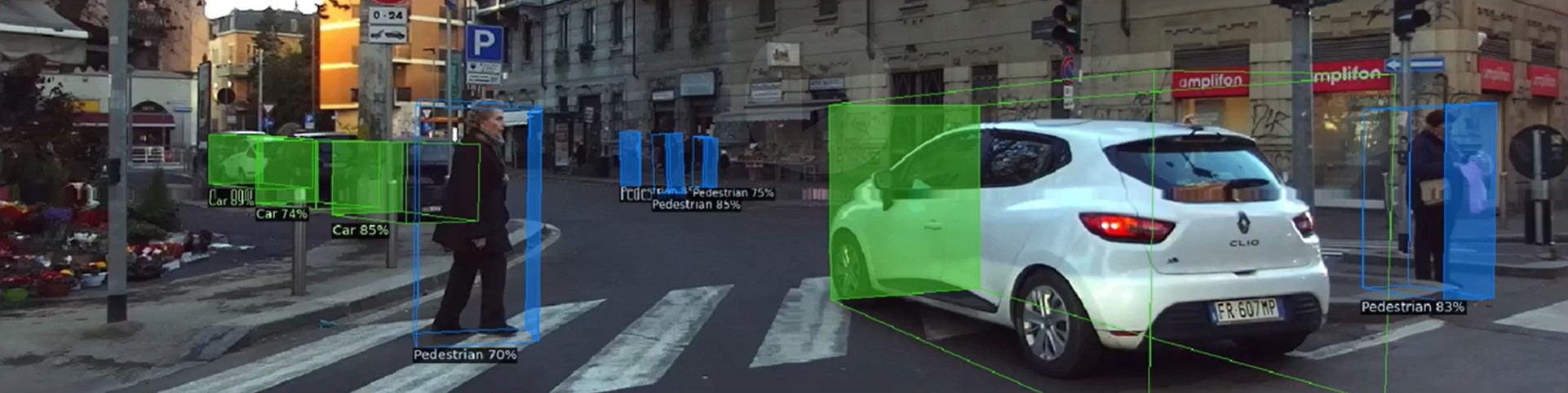}
    \caption{Output of the camera processing module.} 
    \label{classificazione} 
\end{figure}

The aim of the camera processing module is to extract from the image the position and the class of road users in the scene.
Frames are processed using the CNN described in \cite{b12}. We preferred this solution to classical computer vision techniques and other proposed CNN-based methods, since it represented the best compromise between inference time and detection accuracy. An example of the output of this module is shown in Figure \ref{classificazione}.

The output of the module is the set of camera measurements:
\begin{equation}
    \mathcal{Z}_{cam} = \{ \textbf{z}_{cam}^1 , ... , \textbf{z}_{cam}^N \} \; , \; \textbf{z}_{cam}^i = (x_{cam},y_{cam}, \psi_{cam}, c)^T
\end{equation}

where $(x_{cam},y_{cam})$ represent the planar position of the centroid of the detected object with respect to the vehicle, $\psi_{cam}$ is its orientation and $c$ is its class (\textit{e.g.} pedestrian, car, motorbike, ecc.). 

\subsection{LiDAR processing}
The LiDAR processing module extracts from each observation (\textit{i.e.} pointcloud) the position of the centroid of each object visible in the scene.  In this phase, an efficient and robust white box approach was preferred over the use of a 3D detector, being the complexity of the latter solution not necessary for the extraction of the geometrical information required by our MOT approach. The extraction of the LiDAR's measures is performed in the following steps:

\begin{itemize}
    \item \textbf{Ground removal:} points belonging to the ground must be removed to avoid misidentification as obstacles. This step is performed with the approach described in \cite{b13}.
    \item \textbf{Clustering:} in this step, points that belong to the same object are clustered and outliers are removed. The clustering approach used is based on the Euclidean clustering algorithm described in \cite{b14}. 
\end{itemize}

\begin{figure}[h!] 
    \centering 
    \includegraphics[width=0.45\textwidth]{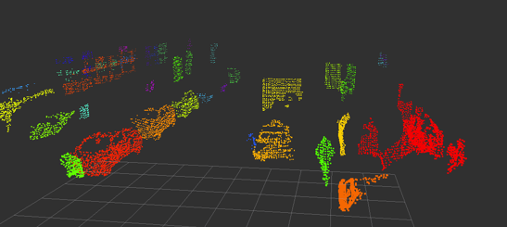}
    \caption{Output example of the LiDAR processing module.} 
    \label{lidar_clustering} 
\end{figure}

An example of this module's output is shown in Figure \ref{lidar_clustering}.

The module outputs the following set of measurements:
\begin{equation}
    \mathcal{Z}_{lidar} = \{ \textbf{z}_{lidar}^1 , ... , \textbf{z}_{lidar}^M \} \; , \; \textbf{z}_{lidar}^i = (x_{lidar},y_{lidar})^T
\end{equation}

where each element represents the planar position of the centroid
of a detected cluster with respect to the vehicle.

\subsection{Data association}
The main goal of the data association module is to merge camera and LiDAR observations that refer to the same object, and to determine if they correspond to an already existing track in the target tracking EKF.
The main hypothesis of the association algorithm is that there is at most one camera or LiDAR measurement for each physical object. Moreover, it is assumed that two distinct objects cannot be associated to the same measurement.
This is achieved for the camera processing output thanks to the 3D detector employed, as it generates a single measurement for every moving object in the scene. The hypothesis is instead not always verified for LiDAR measurements, as the underclustering issue may arise. Underclustering occurs when points belonging to two different objects are wrongly grouped together and only one measurement is returned. In our application, the tuning parameters of the processing algorithm are carefully selected to prevent this behaviour. On the other hand, if the LiDAR processing module produces multiple measurements from the same physical object (overclustering), the association step will handle it by only considering one cluster (the nearest one) and ignoring the others.

Given the sets of measurements $\mathcal{Z}_{lidar}$, $\mathcal{Z}_{cam}$ and the set of tracks $\mathcal{X}$, we iterate the association algorithm three times. 
\begin{enumerate}
    \item \textbf{LiDAR - track association:} in this step, the pairings between existing tracks and new LiDAR measurements are determined. The data association problem is formulated as a typical combinatorial optimization problem known as an association problem, which is solved using Munkres's algorithm \cite{b15}. 
    This requires the construction of a distance matrix $D_{lidar}^{track}$, where each element $d_{ij}$ is computed as follows:
    \begin{equation}\label{eq:distance_lidar_track}
        d_{ij} = (\textbf{z}_{lidar}^i - \textbf{x}_{track}^j)^T S_j (\textbf{z}_{lidar}^i - \textbf{x}_{track}^j)
    \end{equation}
    where $\textbf{z}_{lidar}^i$ is the position of the i-th LiDAR cluster centroid, $\textbf{x}_{track}^j = (x^j, y^j)^T$ is the estimated position of the j-th track and $S_j$ is its corresponding position innovation covariance, which is estimated by the EKF. 
    In addition, Munkres's algorithm considers a gating threshold $\tau_G$ to reject the effect of outliers in the pairing procedure. In particular, every element $d_{ij}$ bigger than $\tau_G$ is forced to infinite, and is therefore ignored by the optimization algorithm.
    Finally, the best measurement - track pairings are determined by solving the optimization problem.
    
   \item \textbf{Camera - track association:} the same association procedure is iterated also for the camera measurements. The distance matrix $D_{cam}^{track}$ is constructed by computing its elements $d_{wj}$ as:
   \begin{equation}\label{eq:distance_camera_track}
        d_{wj} = (\tilde{\textbf{z}}_{cam}^w - \textbf{x}_{track}^j)^T S_j (\tilde{\textbf{z}}_{cam}^w - \textbf{x}_{track}^j)
    \end{equation}
    with $\tilde{\textbf{z}}_{cam}^w = (x_{cam}^w, y_{cam}^w)^T$.
    In this step, the gating distance $\tau_G(c^w)$ is selected depending on the detected class of the w-th camera measurement.
    This is because, due to the characteristics of the 3D detector, the camera position measurement is subject to different errors depending on the type of obstacle.  
    
    \item \textbf{Camera - LiDAR association:} the final step consists in verifying if some close LiDAR and camera measurements exist that are not associated to any track, exploiting again Munkres's algorithm.
    Measurements that have already been paired in the previous iterations are removed from the full sets, and the remaining ones are used to construct the distance matrix $D_{cam}^{lidar}$ constituted by the elements $d_{iw}$:
    \begin{equation}\label{eq:distance_lidar_camera}
        d_{iw} = (\textbf{z}_{lidar}^i - \tilde{\textbf{z}}_{cam}^w)^T \cdot  (\textbf{z}_{lidar}^i - \tilde{\textbf{z}}_{cam}^w)
    \end{equation}
    As for the step (2), $\tau_G(c^w)$ depends on the detected class of the w-th camera measurement. 
\end{enumerate}

After having completed the three association steps the following sets of groups are constructed:
\begin{itemize}
    \item \textbf{Set of LiDAR-track pairs} $(\textbf{z}_{lidar}^i,\textbf{x}^j)$: this set contains the LiDAR measures that are associated to existing tracks, but no corresponding camera measures exists.
    \item \textbf{Set of Camera-track pairs} $(\textbf{z}_{cam}^w,\textbf{x}^j)$: this set contains the camera measures that are associated to existing tracks, but no corresponding LiDAR measures exists.
    \item \textbf{Set of LiDAR-camera-track groups} $(\textbf{z}_{lidar}^i,\textbf{z}_{cam}^w,\textbf{x}^j)$: this set contains the list of tracks that are associated to both a camera and a LiDAR measure.
    \item \textbf{Set of LiDAR-camera pairs} $(\textbf{z}_{lidar}^i,\textbf{z}_{cam}^w)$: this set contains the LiDAR measures that are associated to camera measures, but no corresponding tracks exists.
\end{itemize}

The so formed groups are used to initialize new tracks or correct the motion prediction of existing ones, as further discussed in Sections \ref{sec:meas_function} and \ref{sec:track_management}.

\subsection{Target tracking filter}
At the core of the proposed approach is an Extended Kalman Filter, that predicts the motion of dynamic obstacles through a motion model and corrects the estimate exploiting the measurements coming from the sensor processing modules.

\subsubsection{Motion model}
The tracking filter assumes that the motion of dynamic obstacles follow a CTRV model. The model equations express the relative motion of each tracked object with respect to the ego vehicle. Each track is described by the following state vector: 
\begin{equation}
    \textbf{x} = (x,y,\psi,v, \omega)^T 
\end{equation}
where, $x,y$ are the coordinates of the track relative to the vehicle center of gravity, $\psi$ is its relative heading, and $v,\omega$ its absolute linear and rotational speeds. 
The state equations that describe the evolution of each track, in discrete time, are:
\begin{equation}
\begin{aligned}
    \textbf{x}_{k+1} &=  f(\textbf{x}_k) + \nu_k \\
    \textbf{y}_k &= H\textbf{x}_k + w_k
\end{aligned}
\end{equation}

\begin{equation}
    f(\textbf{x}) = \begin{pmatrix}
    f_1\cdot \cos(\omega^{ego}\cdot T_s) + f_2\cdot \sin(\omega^{ego}\cdot T_s) \\
    -f_1\cdot \sin(\omega^{ego}\cdot T_s) + f_2\cdot \cos(\omega^{ego}\cdot T_s) \\
    \psi + T_s\cdot(\omega - \omega^{ego})\\
    v\\
    \omega 
\end{pmatrix}
\end{equation}
with:
\begin{equation}
    \begin{aligned}
        f_1  &= x + T_s\cdot v\cos(\psi) - T_s\cdot v_x^{ego} \\
        f_2 &= y + T_s\cdot v\sin(\psi) - T_s\cdot v_y^{ego}
    \end{aligned}
\end{equation}
where $T_s$ is the sampling time. Note that, opposite to the classical approach, the CTRV model is written in the body fixed reference frame of the EGO vehicle.

\subsubsection{Measurement function}
\label{sec:meas_function}
Depending on the results of the data association procedure, a different measurement vector may be available for each track at each new iteration.
In particular, three different alternatives are possible:
\begin{itemize}
    \item [-] In the case of a LiDAR-track pair the full measurement vector $\textbf{y}_{lidar}$ is considered:
    \begin{equation}
    \begin{aligned}
    \textbf{y}_{lidar} &= (x_{lidar},y_{lidar})^T \\
    H_{lidar} &= \begin{pmatrix}
        1 & 0 & 0 & 0 & 0\\
        0 & 1 & 0 & 0 & 0
    \end{pmatrix} 
    \end{aligned}
    \end{equation}
    \item [-] In the case of a camera-track pair the full measurement vector $\textbf{y}_{cam}$ is used: 
    \begin{equation}
    \begin{aligned}
    \textbf{y}_{cam} &= (x_{cam},y_{cam}, \psi_{cam})^T \\
    H_{cam} &= \begin{pmatrix}
        1 & 0 & 0 & 0 & 0\\
        0 & 1 & 0 & 0 & 0\\
        0 & 0 & 1 & 0 & 0
    \end{pmatrix} 
    \end{aligned}
    \end{equation}
    \item [-] Finally, in the case of a camera-LiDAR-track group, we consider the position measured by LiDARs and the yaw from the camera. The measurerement matrix $H_{group}$ is equal to $H_{cam}$.
    \begin{equation}
    \begin{aligned}
    \textbf{y}_{group} & = (x_{lidar},y_{lidar}, \psi_{cam})  
    \end{aligned}
    \end{equation}
\end{itemize}

\subsection{Track management}
\label{sec:track_management}
The number of elements tracked by the EKF filter varies dynamically. At each iteration, one of the following situations can happen:
\begin{itemize}
    \item A measurement is available for an existing track. The track state is corrected, using the measurement matrices described in Section \ref{sec:meas_function}.
    \item An existing track is not associated to any measurement. This may happen because the tracked object is shadowed by another obstacle, or because it has exited the scene. In the latter case, the track should be removed.
    \item A measurement is left with no associated track. In this case, a new track should be initialized. To ensure maximum robustness, the track is created only when most reliable information is available, corresponding to the case of a camera-LiDAR measurement. This allows a precise initialization of the position of the new track, thanks to the LiDAR observation, while the detected class from the camera is exploited to easily recognize a potentially dynamic obstacle. LiDAR-only or camera-only measurements are instead not used for initialization, and are therefore discarded.
\end{itemize}

A similar logic to the one described in \cite{b11} is implemented for the initialization and elimination of tracks. When a new camera-LiDAR measurement not linked to a previously existing track is obtained, a new track is initialized as \textit{tentative}. If the new track is associated to at least $M_c$ measurements in the following $N_c$ time instants, it is confirmed, otherwise, it is discarded. A similar approach is also employed for track elimination. A track is removed from the tracking filter if it doesn't receive a minimum of $M_e$ measures within the last $N_e$ time intervals.

\section{Experimental results}

In order to assess the performance and effectiveness of our multi-object tracking approach, we performed two different validations:

\begin{enumerate}[A)]
     \item \textbf{KITTI Multiple Object Tracking (MOT) benchmark}: with this validation it is possible to evaluate the Localization, Detection and Association accuracy of our algorithm.
     \item \textbf{State estimation accuracy:} this validation aims to evaluate the accuracy of the algorithm in estimating the tracked objects position, orientation, speed and yaw rate. The metrics used for the evaluation are the Root Mean Square Error (RMSE), the Mean Absolute Error (MAE) and the Maximum Absolute Error (MaAE). This analysis is performed both in simulation and with real tests.
\end{enumerate}

The vehicle used for the experimental validation is a Maserati MC20, fully equipped with sensors, processing units and actuators for autonomous driving.  
The vehicle is equipped with both proprioceptive and exteroceptive sensors. Proprioceptive sensors are used to compute the linear and rotational velocities of the vehicle, while exteroceptive sensors enable perception of the vehicle's surroundings. The following list details all the sensors mounted on the Maserati MC20 vehicle with their specifications. 

\begin{itemize}
    \item Four wheel encoders.
    \item Two inertial navigation systems (INS): an OXTS AV200 and a Novatel PwrPak7. 
    \item Two Robosense M1 solid state LiDARs, each one guaranteeing an horizontal Field Of View (FOV) of 120$^\circ$ and a vertical one of 25$^\circ$, with a resolution of 0.2$^\circ$. The two LiDARs are positioned with 30$^\circ$ of overlap, in order to obtain a total horizontal FOV of 180$^\circ$.  The two LiDARs are synchronized using the gPTP protocol, allowing for the merging of sensor outputs before the processing step.
    \item A Zed2i camera, with an horizontal FOV of 120$^\circ$ and a focal length of 2.1mm.
\end{itemize}

This vehicle was the first self-driving car to ever participate in the historic 1000 Miglia competition, as part of the 1000MAD project \cite{1000mad}.

\subsection{KITTI MOT benchmark}
This validation uses the KITTI dataset \cite{b16}, known for its practical applicability, diverse object categories, and difficult scenarios, to comprehensively evaluate our multi-object tracking methodology. The metrics used to evaluate the performance of MOT algorithms on the KITTI dataset are described in \cite{b17}. In particular, by focusing on key metrics such as HOTA (Higher Order Tracking Accuracy), DetA (Detection Accuracy), AssA (Association Accuracy), LocA (Localization Accuracy) and MOTA (Multi Object Tracking Accuracy) we can evaluate the algorithm's ability to tackle complex tracking scenarios, including object occlusions, interactions, and varying environmental conditions.
The results are presented in Table \ref{tab:kitti_results} and demonstrate the approach's good performance in tracking both cars and pedestrians.

\begin{table}[htbp]
    \centering
    \begin{tabular}{|>{\centering\arraybackslash}p{1.2cm}|*{5}{>{\centering\arraybackslash}p{1cm}|}}
        \hline
        Class & HOTA (\%) & DetA (\%) & AssA (\%) & LocA (\%) & MOTA (\%) \\
        \hline
        Car & 76.784 & 76.378 & 77.27 & 87.498 & 88.472 \\
        \hline
        Pedestrian & 44.737 & 47.019 & 42.813 & 79.536 & 43.928 \\
        \hline
    \end{tabular}
    \caption{Results of 3D MOT on the KITTI tracking validation set for the car and pedestrian classes.}
    \label{tab:kitti_results}
\end{table}

\subsection{State estimation accuracy}

This analysis is performed considering both a simulated and a real scenario. As regards the simulation, we exploited VI-WorldSim, an advanced simulator for autonomous driving that allows the creation of detailed urban scenarios populated with different types of agents. The simulator also faithfully replicates the vehicle's sensor setup and observations.
For the validation, we created a dedicated scenario wherein the vehicle drives on city streets alongside other agents, such as pedestrians, cyclists and other vehicles. 
The validation results are detailed Table \ref{tab:agent_metrics}. The RMSE, MAE and MaAE remain generally small for all agent categories (e.g. car, cyclist, pedestrian). The position estimation error of the car class is slightly higher compared to cyclists and pedestrians, but the maximum errors remain less than 1 m. On the contrary, the estimation of other state variables shows better results for the car class than for pedestrians and cyclists, particularly in terms of orientation. This outcome may be explained by the better performance of the 3D detector used to measure the orientation of cars than of cyclists and pedestrians. Furthermore, it is crucial to note that pedestrians and cyclists change their orientation more rapidly than cars, leading to higher errors in orientation estimates for these categories. This validation demonstrates the algorithm's ability to accurately and reliably track the position, orientation, and dynamics of objects.

\begin{table*}[t]
    \centering
    \begin{tabular}{|l|c|c|c|c|c|c|c|c|c|c|c|c|}
        \hline
        \multirow{2}{*}{Agent} & \multicolumn{4}{|c|}{RMSE} & \multicolumn{4}{|c|}{MAE} & \multicolumn{4}{|c|}{MaAE} \\
        \cline{2-13}
         & pos [$m$] & $\psi$ [$^\circ$] & $v$ [$m/s$] & $\omega$ [$^\circ/s$] & pos [$m$] & $\psi$ [$^\circ$] & $v$ [$m/s$] & $\omega$ [$^\circ/s$] & pos [$m$] & $\psi$ [$^\circ$] & $v$ [$m/s$] & $\omega$ [$^\circ/s$] \\
        \hline
        \hline
        Cyclist & 0.253 & 13.34 & 0.334 & 9.386 & 0.207 & 7.351 & 0.174 & 6.019 & 0.697 & 31.87 & 1.157 & 35.15 \\
        \hline
        Car 1 & 0.516 & 5.946 & 0.574 & 7.936 & 0.406 & 3.391 & 0.438 & 5.767 & 0.955 & 17.38 & 1.622 & 31.31 \\
        \hline
        Car 2 & 0.408 & 5.123 & 0.498 & 7.241 & 0.312 & 2.954 & 0.354 & 4.965 & 0.875 & 15.24 & 1.421 & 24.12 \\
        \hline
        Car 3 & 0.492 & 5.561 & 0.524 & 7.532 & 0.361 & 3.465 & 0.398 & 5.482 & 0.894 & 16.54 & 1.574 & 28.63 \\
        \hline
        Pedestrian 1 & 0.143 & 17.79 & 0.184 & 11.32 & 0.107 & 12.04 & 0.137 & 7.521 & 0.379 & 47.16 & 0.669 & 37.47 \\
        \hline
        Pedestrian 2 & 0.158 & 18.27 & 0.214 & 11.86 & 0.112 & 12.47 & 0.158 & 7.864 & 0.385 & 53.14 & 0.884 & 40.35 \\
        \hline
        Pedestrian 3 & 0.167 & 15.39 & 0.193 & 9.945 & 0.132 & 10.58 & 0.125 & 6.361 & 0.401 & 41.37 & 0.563 & 32.75 \\
        \hline
    \end{tabular}
    \caption{MOT algorithm's state estimation errors for different agents (simulated scenario).}
    \label{tab:agent_metrics}
\end{table*}

\begin{figure}[h!]
    \centering 
    \includegraphics[width=0.45\textwidth]{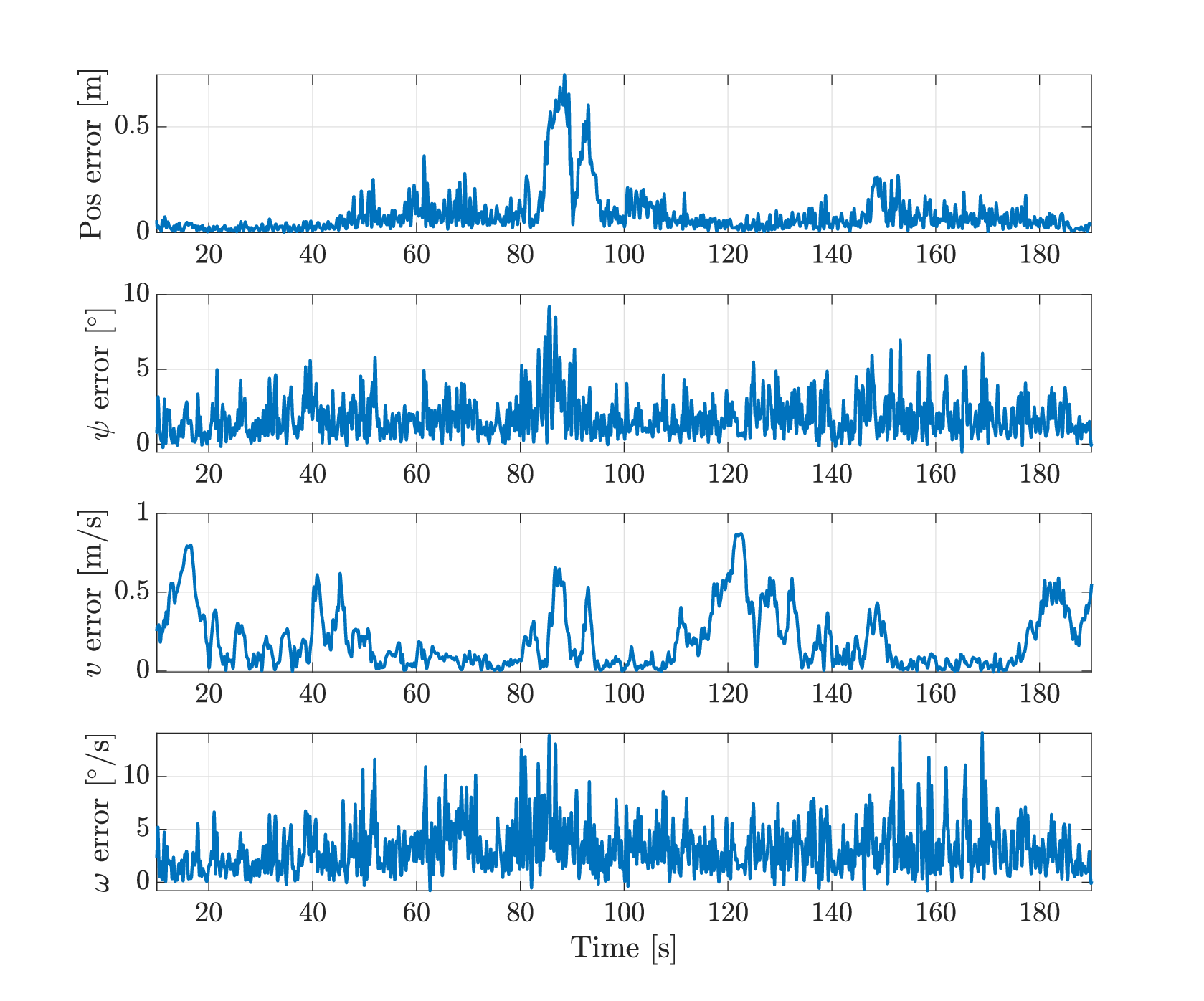}
    \caption{Track state estimation error. The four plots show respectively the position, orientation, speed and yaw rate estimate errors.} 
    \label{fig:exp_test_errors} 
\end{figure}

The proposed algorithm was validated also with real-world tests. For this validation a chosen vehicle was equipped with an  GNSS/INS unit with RTK correction that generates ground-truth measurements. 
The results of this experimental test, displayed in Figure \ref{fig:exp_test_errors}, demonstrate the significant level of performance achieved by the proposed algorithm. The algorithm consistently achieves low levels of estimation errors for the position, orientation, speed, and yaw rate of the tracked vehicle, which align with the error values obtained during simulation. 

Moreover, the experimental validation results were compared with the same algorithm operating in single-modality modes, exclusively using either camera or LiDAR data. The results are presented in Table \ref{tab:errors_comparison}. Notably, the algorithm yielded lower performance in both the \emph{camera-only} and \emph{LiDAR-only} modes, underscoring the necessity of sensor fusion to achieve accurate tracking by combining information from both sensors.

\begin{table}[htbp]
    \centering
    \begin{tabular}{|l|l|>{\centering\arraybackslash}p{1.55cm}|>{\centering\arraybackslash}p{1.55cm}|>{\centering\arraybackslash}p{2.0cm}|}
        \hline
         \multicolumn{2}{|c|}{} & Single-modal (camera) & Single-modal (LiDAR) & Multi-modal \hspace{0.5cm} (camera+LiDAR)\\
        \hline
        \hline
        \multirow{4}{*}{RMSE} & pos [$m$] & 0.1804 & 0.1671 & 0.1378 \\
        \cline{2-5}
        & $\psi$ [$^\circ$] & 2.171 & 3.035 & 2.329 \\
        \cline{2-5}
        & $v$ [$m/s$] & 0.2782 & 0.2811 & 0.273 \\
        \cline{2-5}
        & $\omega$ [$^\circ/s$] & 3.674 & 4.517 & 4.15 \\
        \cline{2-5}
        \hline
        \hline
        \multirow{4}{*}{MAE} & pos [$m$] & 0.1271 & 0.097 & 0.085 \\
        \cline{2-5}
        & $\psi$ [$\circ$] & 1.663 & 1.877 & 1.786 \\
        \cline{2-5}
        & $v$ [$m/s$] & 0.2016 & 0.201 & 0.196 \\
        \cline{2-5}
        & $\omega$ [$^\circ/s$] & 2.809 & 3.286 & 3.185 \\
        \cline{2-5}
        \hline
        \hline
        \multirow{4}{*}{MaAE} & pos [$m$] & 1.015 & 0.845 & 0.797 \\
        \cline{2-5}
        & $\psi$ [$^\circ$] & 11.017 & 31.939 & 9.614 \\
        \cline{2-5}
        & $v$ [$m/s$] & 0.8787 & 1.238 & 0.873 \\
        \cline{2-5}
        & $\omega$ [$^\circ/s$] & 16.227 & 22.803 & 14.744 \\
        \cline{2-5}
        \hline
    \end{tabular}
    \caption{Errors comparison for the three different modalities.}
    \label{tab:errors_comparison}
\end{table}

\section{Conclusions}
This paper presents a MOT algorithm that fuses camera and LiDAR sensors. The method utilizes a camera 3D detector to detect dynamic obstacles and clustering techniques to process the LiDAR output, ensuring fast and precise object positioning. 
Our MOT algorithm tracks each object using an EKF and a novel motion model that estimates the position, orientation and absolute longitudinal and angular velocities of the object, without relying on maps or knowledge of the ego global pose. This approach only requires the current measured position and orientation of the observed object, as well as the longitudinal and angular velocities of the ego vehicle. The correction phase measurement vector and matrix are dynamically adapted based on the results of a three-step data association procedure that groups the tracks and measurements more likely to have been generated by the same object. 


\vspace{12pt}

\end{document}